\documentclass[10pt,conference]{IEEEtran}

\usepackage{amssymb}
\usepackage{amsfonts}
\usepackage{amsmath}
\usepackage{bbding}
\usepackage{color}
\usepackage{comment}
\usepackage{epsfig} 
\usepackage{graphics}
\usepackage{lipsum}
\usepackage{subfigure}
\usepackage{multirow}
\usepackage{times}
\usepackage{tcolorbox}
\usepackage{url}
\usepackage{wrapfig}
\usepackage{svg}
\usepackage[ruled,vlined]{algorithm2e}

\IEEEoverridecommandlockouts

\newcommand{\eat}[1]{}

\begin{document}
\title{PPG-based Heart Rate Estimation with Efficient Sensor Sampling and Learning Models\thanks{This project was funded in part by The University of Texas at San Antonio, Office of the Vice President for Research, Economic Development \& Knowledge Enterprise. This work appeared at the 2022 IEEE International Conference on Embedded Software and Systems (ICESS)}}

\author{
\IEEEauthorblockN{Yuntong Zhang$^1$, Jingye Xu$^1$, Mimi Xie$^1$, Wei Wang$^1$, Keying Ye$^2$, Jing Wang$^3$ and Dakai Zhu$^1$}\\
\IEEEauthorblockA{$^1$Department of Computer Science, The University of Texas at San Antonio, USA\\
$^2$Department of Management Science and Statistics, The University of Texas at San Antonio, USA\\
$^3$College of Nursing, Florida State University, USA
}}


\maketitle
  
\thispagestyle{empty} 
\pagestyle{empty}

\begin{abstract}

Recent studies showed that Photoplethysmography (PPG) sensors embedded in wearable devices can estimate heart rate (HR) with high accuracy. However, despite of prior research efforts, applying PPG sensor based HR estimation to embedded devices still faces challenges due to the energy-intensive high-frequency PPG sampling and the resource-intensive machine-learning models. In this work, we aim to explore HR estimation techniques that are more suitable for lower-power and resource-constrained embedded devices. More specifically, we seek to design techniques that could provide high-accuracy HR estimation with low-frequency PPG sampling, small model size, and fast inference time. First, we show that by combining signal processing and ML, it is possible to reduce the PPG sampling frequency from 125 Hz to only 25 Hz while providing higher HR estimation accuracy. This combination also helps to reduce the ML model feature size, leading to smaller models. Additionally, we present a comprehensive analysis on different ML models and feature sizes to compare their accuracy, model size, and inference time. The models explored include Decision Tree (DT), Random Forest (RF), K-nearest neighbor (KNN), Support vector machines (SVM), and Multi-layer perceptron (MLP). Experiments were conducted using both a widely-utilized dataset and our self-collected dataset. The experimental results show that our method by combining signal processing and ML had only $5\%$ error for HR estimation using  low-frequency PPG data. Moreover, our analysis showed that DT models with 10 to 20 input features usually have good accuracy, while are several magnitude smaller in model sizes and faster in inference time.

\end{abstract}

\section{Introduction}

\eat{HR are critical health vital signs and it is necessary to monitor them for different purposes continuously; 
the traditional approach to obtaining HR is to adopt ECG devices, which are hard to operate and inconvenient to utilize; 
Several wearable devices (such as chest straps, and smartwatches) have been designed to monitor HR using PPG sensors. However, most such devices do not provide an interface to access data for further analysis. 
In this work, with the objective of developing a low-power and sustainable wearable device to monitor HR using PPG sensors, we study various ML models and data processing methods to obtain accurate HR from PPG sensor data. Explain data processing methodology and ML models, and summary of results; 
list of contributions; paper organization
}

Heart rate (HR) is an important vital sign for the cardiovascular system and has been widely used as a biomarker for diagnostic and early prognostic of several diseases such as hypertension and heart failure~\cite{fox2007resting}. Besides the critical condition monitoring in the hospital setting, many applications also depend on continuously measured HR, such as fitness tracking, bio-metric identification, and frailty detection~\cite{biswas2019cornet,biswas2019heart,eskandari2022frailty}. Therefore, it is desirable to have a real-time HR monitoring system, which can conveniently provide accurate data in an effective manner to support such applications. 

The traditional and reliable approach to continuously monitoring HR is to utilize electrocardiogram (ECG) devices, which are generally expensive and inconvenient to deploy for outpatients or other users to operate in a continuous manner. Besides ECG, Photoplethysmography (PPG) sensors, when applied to the surface of the skin, like fingers, wrist, or earlobe, can utilize light signal to monitor changes in blood flow, which can be exploited to derive HR~\cite{castaneda2018review}. Given its low cost and convenience, PPG sensors have been widely utilized as an inexpensive alternative to monitor HR in wearable embedded devices (e.g., smartwatches), which have limited energy and computing resources. However, as PPG is susceptible to motion artifacts (MA), existing PPG-based work that removes MA to detect HR typically has two limitations.



First, the classical method for extracting HR from PPG data is based on signal processing, which usually requires a relatively high sampling frequency (hundreds of Hz) and a complex process to remove MA noises to achieve high accuracy. This high sampling frequency may incur high energy consumption, preventing these signal processing techniques from being applied to energy and resource-constrained devices. 
For example, a widely utilized data set for HR estimation is the IEEE Signal Processing Cup (ISPC) data set, which was proposed by Zhang et al., along with a signal processing-based algorithm that includes signal decomposition, sparse signal reconstruction, and spectrum peak tracking, to extract HR from PPG data~\cite{zhang2014troika}.
The ISPC dataset contains PPG signals sampled at a high frequency of 125 Hz, which was widely adopted by other studies~\cite{zhang2015photoplethysmography,bashar2019machine,puranik2019heart,chang2021deepheart}.
However, Bhowmik et al. has reported that the PPG sensor on smartwatch sampling at 100 Hz consumed significantly more energy than 25 Hz~\cite{bhowmik2017novel}.
Furthermore, certain signal processing techniques may require intricate algorithms to achieve high accuracy, which further worsens energy consumption~\cite{tobola2015sampling}. 

Second, with the advancement of machine learning (ML), several studies exploited various machine learning models to remove MA noises and estimate HR.
However, ML models, especially neural networks, usually are computationally intensive, limiting their application to embedded devices with limited resources. 
For example, Wittenberg et al. used complex deep learning models including convolutional neural network (CNN) and Gated recurrent unit (GRU) for PPG peaks detection \cite{wittenberg2020evaluation}. 
There were also studies that employed ML models such as K-means, Random Forest (RF), and Bayesian learning algorithm~\cite{bashar2019machine,alqaraawi2016heart}. 
Moreover, existing studies employ ML models~\cite{everson2019biotranslator,biswas2019cornet,xu2019deep} usually employed a large number of features (i.e., PPG signals) to achieve high accuracy, which led to large models that may not fit in small embedded devices.
To enable ML-based HR monitoring in a resource-constrained embedded device, there needs research to investigate feature dimension reduction to allow smaller ML models. There also needs exploration to determine the type of ML models that can provide accurate HR readings with less resource usage. 
In this paper, we report the design of our HR monitoring solution using an off-the-shelf PPG sensor. The system is specially designed for resource-constrained devices and addresses the above limitations. To address the limitations, we combine signal processing and ML methods. That is, we first applied signal processing to generate rough HR estimations. Then these rough HRs are passed through a smaller ML model to generate more accurate HR estimations.
On one hand, applying the ML model to signal-processing-generated HRs allowed us to sample PPG signals at only 25 Hz while achieving higher accuracy because ML models can be trained specifically to improve accuracy and remove MA with low-frequency samples. On the other hand, applying signal processing before ML eliminates the need for the ML model to directly take large numbers of PPG signals as inputs, reducing feature sizes and model sizes. 

To address the ML model type issue of the second limitation, we compare the accuracy, model size, and inference time of five different ML models, including Decision Tree (DT), Random Forest (RF), K-nearest neighbor (KNN), Support Vector Machines (SVM) and Multi-layer Perceptron (MLP), and provide insights on which are more suitable for the resource-constrained device. Moreover, because existing data sets are usually too short for extensive evaluation, we also collected a new data set for HR study which is long enough for HR ML model training and testing with different features.

Our experimental evaluations show that the system can achieve less than 5\% mean average prediction errors (MAPE) for HR estimation with a PPG sampling rate of only 25 Hz. Moreover, our exploration results show that Decision Tree (DT) models usually could provide accurate estimation with a smaller model size of about 10 KB and a shorter inference time of less than 3 microseconds ($\mu$s).

The contributions of this paper include:

\begin{itemize}
\item Novel HR estimation methodologies that combined both signal processing and ML models, allowing high accuracy HR estimations with low-frequency PPG signals, fewer ML features, and smaller ML models. 

\item A systematic analysis of different ML model types and feature sizes to study their impact on the accuracy in HR estimation. This analysis showed that all considered ML models can provide about 5\% MAPE for both the ISPC dataset and our collected dataset. 

\item Comprehensive evaluations of the model size and inference time for different ML model types and configurations to study their suitability for the resource-constrained HR monitoring environment. This analysis found that DT can provide a good balance between accuracy, model size, and inference time.
\end{itemize}

The rest of the paper is structured as follows.
Section~\ref{sec:related_work} discusses the related work. 
The methodology and implementation are detailed in Section~\ref{sec:method}. 
Evaluation results are presented in Section~\ref{sec:Evaluation} and the conclusions are drawn in Section~\ref{sec:Conclusion}.

\section{Background and Closely Related Work}
\label{sec:related_work}

In this section, we present the background and closely related work on HR estimation.

\subsection{HR from ECG}

The traditional medical device to measure HR is ECG.
ECG records heart activity utilizing electrodes placed at certain skin spots on the human body and produces an electrocardiogram, which is a graph that shows the heart's electrical activity over time.
An electrocardiogram contains the QRS complexes information, which is the most important waveform in an electrocardiogram that shows the spread of a stimulus through the ventricles~\cite{dohare2014efficient,GOLDBERGER201811}. 
RR intervals can be derived from the QRS complex, which in turn, gives HR. 
More specifically, because the RR interval is the interval between heartbeats, the reciprocal of the RR interval is the HR~\cite{LANFRANCHI2011226}.
Although ECG can produce accurate HR, attaching electrodes to the human body makes it inconvenient to use.

\subsection{HR from PPG - Signal Processing}

Due to their convenient usage and small sizes, Photoplethysmography (PPG) sensors have become a popular replacement for ECG in HR monitoring.
PPG uses light signals to monitor blood flow.
Heartbeats cause periodical changes in the blood flow, which cause periodical changes in the reflected light received by the PPG sensor.
Hence, periodical PPG light wave changes can be exploited to derive the heartbeat.

The main issue with PPG sensors is the noise in the signal, which is usually the result of motion artifacts (MA).
The common method to remove MA and calculate HR is signal processing, which typically tracks the peaks of the PPG signals. Zhang et al. proposed an algorithm that consists of signal decomposition, sparse signal reconstruction, and spectrum peak tracking to extract HR from PPG signals in intensive physical activities environment~\cite{zhang2014troika}.
Their dataset, denoted as the IEEE Signal Processing Cup (ISPC) 2015 dataset, has been widely utilized for evaluating HR monitoring solutions. 
It includes PPG sensor data, accelerometer sensor data, and ECG data. 
In this ISPC dataset, the PPG sensor is installed on the wrist and the ground truth is a one-channel ECG.
In contrast, we applied PPG on fingertips and utilized a three-lead ECG to get ground truth HR. Moreover, each ISPC data recording lasts for only a few minutes, whereas each trace in our data lasts for about two hours.

\subsection{HR from PPG - Machine Learning}

Recently, ML has been found to be a promising method to remove MA from PPG signals and estimate HR.
Prior studies utilized various ML algorithms to estimate HR. 
For example, Bashar et al. employed K-means clustering and Random Forest (RF) for HR estimation~\cite{bashar2019machine}.
The raw PPG signal was pre-processed with a 2nd order bandpass filter.
Then the K-means clustering algorithm was used to identify noisy data and Random Forest regression was used to predict HR based on PPG and acceleration data.
They used features extracted from PPG signals and also examined adding features from accelerometers.
Puranik and Morales estimated HR in real-time with an adaptive Neural Network filter and a post-processing smoothing and median filter~\cite{puranik2019heart}. 
Chang et al. came up with DeepHeart, an HR estimation approach that combines deep learning and spectrum analysis~\cite{chang2021deepheart}.
The raw PPG signal was pre-processed by a three-order Butterworth bandpass filter. 
Then the PPG signal was sent to an ensemble model to remove noise.
The ensemble model contains several deep learning models with convolutional layers.
Biswas et al. proposed CorNET, a convolutional neural network with long short-term memory (LSTM) to estimate HR and perform biometric identification based on PPG signals~\cite{biswas2019cornet}.
The raw PPG signal was pre-processed with a band-pass 4th order Butterworth filter and a normalizer.
This method was estimated with leave-one-window-out validation on the ISPC dataset.
Rocha et al. designed Binary CorNET, a binarized CorNET to estimate HR 
~\cite{rocha2020binary}.
All the above works used the ISPC dataset in their evaluation, which shows the popularity of the ISPC dataset.
However, as pointed out in \cite{reiss2018ppg}, the ISPC dataset is insufficient for deep learning approaches since the available data for different activities is short and the total number of samples is limited. The ISPC dataset also used a high (125 Hz) sampling frequency, which incurs high energy usage.

Reiss et al. presented a CNN architecture for PPG-based HR estimation \cite{reiss2018ppg}. They first preprocessed the PPG sensor data with FFT and z-normalization, and then train and evaluate the CNN model using leave-one-session-out (LOSO) cross-validation with the ISPC dataset.
In the experiment, they compared the proposed method with three classical signal processing methods and concluded that the performance of the CNN-based HR estimation is comparable with classical methods.
Panwar et al. designed a convolutional neural network with LSTM to estimate HR and blood pressure based on PPG signals \cite{panwar2020pp}. They evaluated the model on the MIMIC dataset \cite{saeed2011multiparameter} and showed the effectiveness of the neural model. 
The MIMIC dataset is collected from patients, while in this work, we focus on HR for healthy subjects.

Unlike prior studies that applied signal processing only for data pre-processing, our solution used signal processing first to generate a set of HR estimations, which are then further processed by ML models to improve estimation accuracy. By combining signal processing based and ML-based HR estimation, we can reduce both PPG sampling frequency and ML feature size while retaining high accuracy.

\section{Learning-Oriented Efficient HR Estimation with PPG Data}
\label{sec:method}

In this section, we present the methodology for learning-oriented efficient HR estimation that combines signal processing and ML models using PPG data.


\subsection{PPG-based HR monitoring system}\label{sec:ppg_system}

\begin{figure}
  \centering
  \includegraphics[width=1\linewidth]{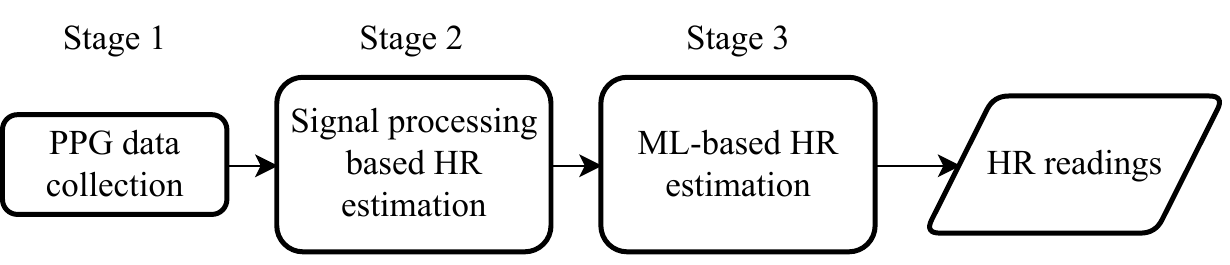}
  \caption{The architecture of the PPG-based HR monitoring system.}
  \label{fig:system}
\end{figure}

Fig.\ref{fig:system} shows the architecture of our HR monitoring system, which has four stages.
In Stage 1, data collection modules collect the PPG signals. Here, a PPG sensor is attached to the subject's fingertip, which outputs red and infrared light signals. In Stage 2, the signal processing module reduces noises in PPG data and generates rough estimations of HR for every second.
In Stage 3, an ML module takes a sequence of rough HRs to get more accurate HR estimations.
Finally, a report module reports the estimated HR readings to users. 
The rest of this section provides a detailed description of Stages 2 and 3.

\subsection{HR Estimation}

\begin{figure}
  \centering
  \includegraphics[width=1\linewidth]{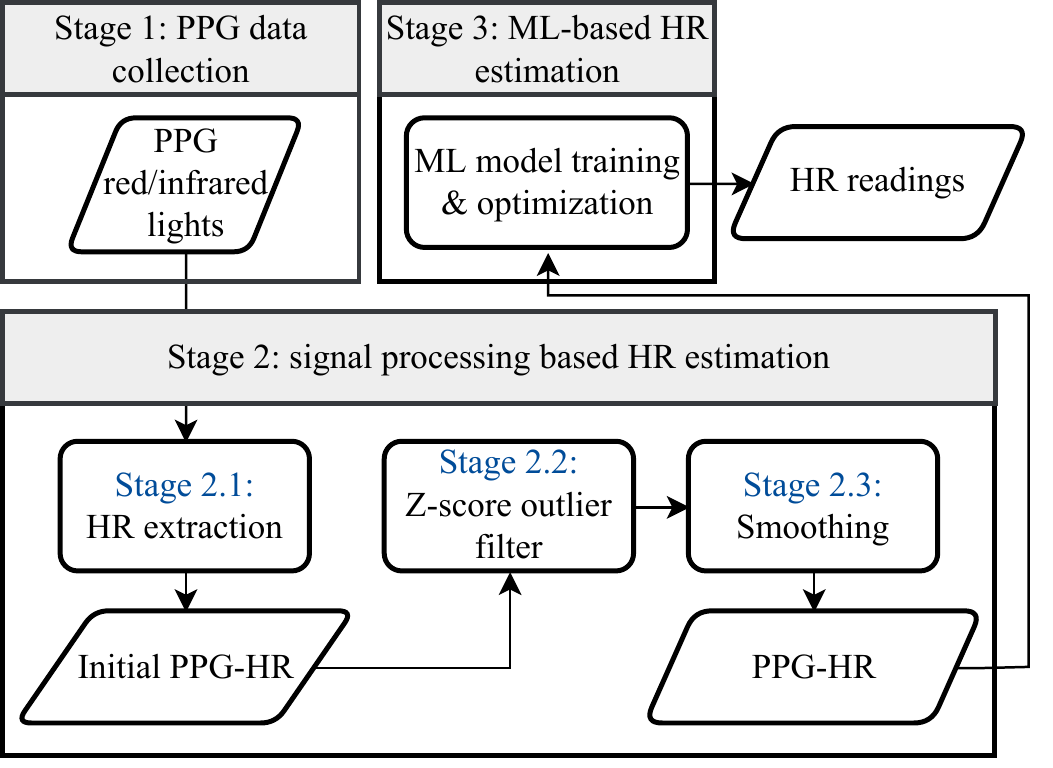}
  \caption{Major steps of the PPG-based HR monitoring system.}
  \label{fig:stage2}
\end{figure}

Figure~\ref{fig:stage2} provides the detailed processing steps of our HR monitoring system, which are described as follows.

\subsubsection{Signal Processing based HR Estimation (Stage 2)}\label{sec:stage2}
The input data for Stage 2 are time-series signals generated by PPG in the PPG data collection stage, which contains the red/infrared light signals reflected by blood under the skin. 
We employed a sampling rate of 25 Hz, i.e., 25 signal samples every second.

\paragraph{Initial HR extraction (Stage 2.1)}
The processing in stage 2.1 takes the PPG light signals from the PPG data collection stage, finds the local peaks within the light sequence, and then calculated by counting the number of peaks to estimate HR~\cite{ppg_allen2007photoplethysmography}. Currently, our signal processing algorithm converts 25 signals in a second to 4 HR readings for that second (denoted as initial PPG-HR).

\paragraph{Z-score outlier filter (Stage 2.2)}
Due to motion artifacts, there may be large noises in PPG signals, leading to outliers in the initial HR estimations from stage 2.1. To eliminate these outliers, a Z-score outlier filter is applied.

The Z-score filter is a popular method used to find outliers. Given a data point, its z-score represents the distance (i.e., deviation) between its value and the mean of all data points, measured by multiples of standard deviation ($std$)~\cite{zscore1_mendenhall2016statistics, zscore2_spiegel2018schaum}. For example, if a data point's z-score is 3, then the difference between this data point and the mean is $3\times std$.
Therefore, if the z-score (i.e., distance) is larger than a threshold, then the data point can be viewed as an outlier. 
Here, we set the threshold z-score to 3 following common practice~\cite{zscore3_zhang2011illustration, zscore4_vysochanskij1980justification}. 

However, we cannot simply remove the outliers because HR is a time series that will be utilized as features by ML models. Here, deleting any values will possibly lose the time-related information. Therefore, we choose to revise the outliers' value to be the average of their two surrounding HR readings.

\paragraph{Smoothing (Stage 2.3)}
Even after the Z-score filter, there still could be abrupt peak or valley HR readings that are overly higher or lower than their surrounding HRs due to PPG signal noises,
Therefore, to further reduce the PPG HR fluctuation, we applied additional data smoothing. 

To smooth, we first take an average of the four HR readings within a second to convert them into one HR per second. This smoothing removes fluctuations of the HRs within a second. It also reduces the number of HRs to be input to the ML models in Stage 3.

Moreover, the HR reading of a person in general does not change abruptly in one second and returns back in the next second. That is, an HR reading should not be significantly different from the HRs before and after it. Hence, it is possible to smooth the HR based on a specific upper and lower boundary to further restrict the PPG HR fluctuation. In our experiment, we set the boundary to 5\%. That is, an HR reading can only be less than or equal to 5\% above or below its predecessor.
For HR readings that are more than 5\% higher (or lower) than the previous value, their values are changed to be 5\% higher (or lower) than the previous value. This 5\% boundary is determined based on the fluctuation range observed from the more reliable ECG data.

\subsubsection{ML-based HR Estimation (Stage 3)}\label{sec:ml_estimation}
The rough HR estimations generated by signal processing (denoted as PPG-HR) in Stage 2 are passed to an ML model to generate a more accurate HR estimation. As our PPG sensor sampling rate was set at 25Hz for power consideration to preserve energy, the PPG-HRs from these samples may still contain large errors. The ML model is particularly trained to reduce errors due to the low sampling frequency. 

More specifically, the ML model takes a sequence of the last $k$ PPG-HR estimations to generate the current HR reading. Intuitively, the ML model uses these $k$ PPG-HRs to assess the potential errors in them to produce a more accurate HR reading. We evaluated different values for $k$ in our experiments. We also evaluated different types of ML models, including DT, RF, KNN, SVM, and MLP. These evaluation results are reported in Section~\ref{sec:Evaluation}.

\subsection{Data Sets and Model Training}
\subsubsection{Our Data Set and Model Training}
As discussed previously, the features of our ML models are $k$ last PPG-HR estimations, i.e., the PPG-HR estimations from the last $k$ seconds given one PPG-HR per second. The labels (i.e., ground truth HR) are obtained by ECG. During data collection, our subject wears the PPG monitoring system (described in Section~\ref{sec:ppg_system} and ECG electrodes at the same time for data recording. Hence, the ECG can provide ground truth HR for corresponding PPG signals and PPG-HR. The subject engaged in three scenarios: sitting, sleeping, and conducting daily activities. The daily activity includes office working, walking, drinking water, etc. Each recording lasted for 2 hours.

The collected data sets are then partitioned into training and testing data sets, with a split of 80\% and 20\%, respectively. All ML models also went through random search based hyperparameter tuning~\cite{bergstra2012random} to find the best model.

\subsubsection{ISPC Data Sets and Model Training}
To show that the accuracy benefit from our combined signal processing and ML method is generic, we also applied our method to the popular ISPC dataset~\cite{zhang2014troika}. This dataset provides raw PPG signals at 125 Hz along with ground truth HR readings. 

Note that, since ISPC data sets used a different PPG sensor and different sampling rate, our signal processing steps cannot be applied to it. Therefore, we applied the signal processing steps in the original paper of the ISPC data set~\cite{zhang2014troika}. Moreover, because the source code of the ISPC data set's signal processing is not publicly available, and the potential changes of the dataset after its publication, the HR estimation errors from our reproduced signal processing are not completely the same as those reported in the original paper.

\section{Experimental Evaluation}
\label{sec:Evaluation}


\subsection{Experiment Setup}


\subsubsection{ISPC dataset}
The ISPC data set contains one channel of ECG data, 2 channels of PPG data, and 3-axis acceleration data. There are 12 healthy subjects and each of them is required to do a series of exercises to generate 5-minute long recordings with a PPG sampling rate of 125 Hz. Although there are two channels of PPG signals, we only used one channel that has lower noise for HR estimation for better reproduction accuracy. 80\% of the data are used for training, and 20\% are used for testing.

\subsubsection{Our dataset}

\begin{figure}
  \centering
  \includegraphics[width=0.7\linewidth]{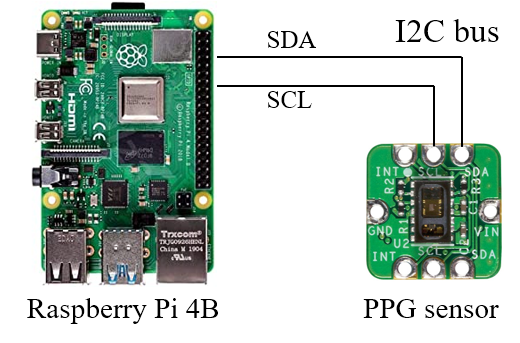}
  \caption{Connections between Raspberry Pi and PPG sensor.}
  \label{fig:pi}
\end{figure}

Our data set contains one channel of PPG data and 3-lead ECG signals.
The subject simultaneously wore the PPG sensor on the fingertip and ECG electrodes on the torso. The PPG sensor is connected to a Raspberry Pi 4B through an I2C bus to record the data, as shown in Fig.\ref{fig:pi}.
The hardware components include: 1) a Raspberry Pi 4B with 4 GB RAM; 2) a Maxim MAXREFDES117\footnote{https://a.co/d/1VQGxm8} HR monitor with a MAX30102 PPG sensor; and 3) a TLC5007 Dynamic ECG (not shown in the figure).



\subsubsection{ML Configurations}
ML algorithms are implemented using Scikit-Learn~\cite{scikit-learn}.
We used the random search function {\em RandomizedSearchCV} in Scikit-Learn to tune the model hyperparameters. The hyperparameters are: 
1) For DT, the maximum depth of the tree ranges from 1 to 20. 
2) For RF, the number of trees is between 1 to 30, and the maximum tree depth is between 3 to 7. 
3) For KNN, the number of neighbors is between 1 to 30, and the distance can be Manhattan or Euclidean. 
4) For SVM, the kernel may be among RBF, sigmoid and polynomial, and regularization (i.e., $C$) may be between 0.00001 to 10. 
5) For MLP, there are 3 hidden layers with 2 to 15 neurons in each layer.
The activation function can be {\em relu} or {\em tanh}. 
The L2 regularization hyper-parameter $\alpha$ search range is from 0.00001 to 10.


\begin{table*}
\caption{MAPE and Standard Deviation (SD) for different number of features when estimating ISPC HR (MAPE±SD).}
\centering
\renewcommand{\arraystretch}{1.5}
\begin{tabular}{|c|c|c|c|c|c|c|}
\hline
\multicolumn{1}{|l|}{\multirow{2}{*}{}}& \multicolumn{6}{c|}{Number of features}  \\ \hline
\multicolumn{1}{|l|}{} & 2             & 4             & 6             & 8             & 10            & 15            \\ \hline\
DT                     & 3.86\%±5.26\% & 3.77\%±4.91\% & 3.58\%±4.14\% & 3.4\%±3.64\%  & 3.41\%±3.53\% & \textbf{2.76\%±1.89\%} \\ \hline
RF                     & 3.14\%±3.45\% & 3.29\%±3.4\%  & 2.9\%±2.37\%  & 2.79\%±2.08\% & 2.7\%±1.87\%  & \textbf{2.62\%±1.91\%} \\ \hline
KNN                    & 2.89\%±2.6\%  & 2.89\%±2.2\%  & 2.9\%±1.87\%  & 3.08\%±1.75\% & 3.55\%±2.33\% & 3.69\%±1.94\% \\ \hline
SVM                    & 3.81\%±2.5\%  & 5.14\%±6.18\% & 4.71\%±5.49\% & 3.83\%±2.2\%  & 5.11\%±5.41\% & 4.06\%±2.49\% \\ \hline
MLP                    & 3.33\%±5.86\% & 3.26\%±5.2\%  & 3.38\%±5.49\% & 4.53\%±5.78\% & 3.4\%±4.41\%  & 4.81\%±4.46\% \\ \hline
\end{tabular}
\label{tab:hr_numofFeature_ispc}
\end{table*}

\begin{figure}
  \centering
  \includegraphics[width=1\linewidth]{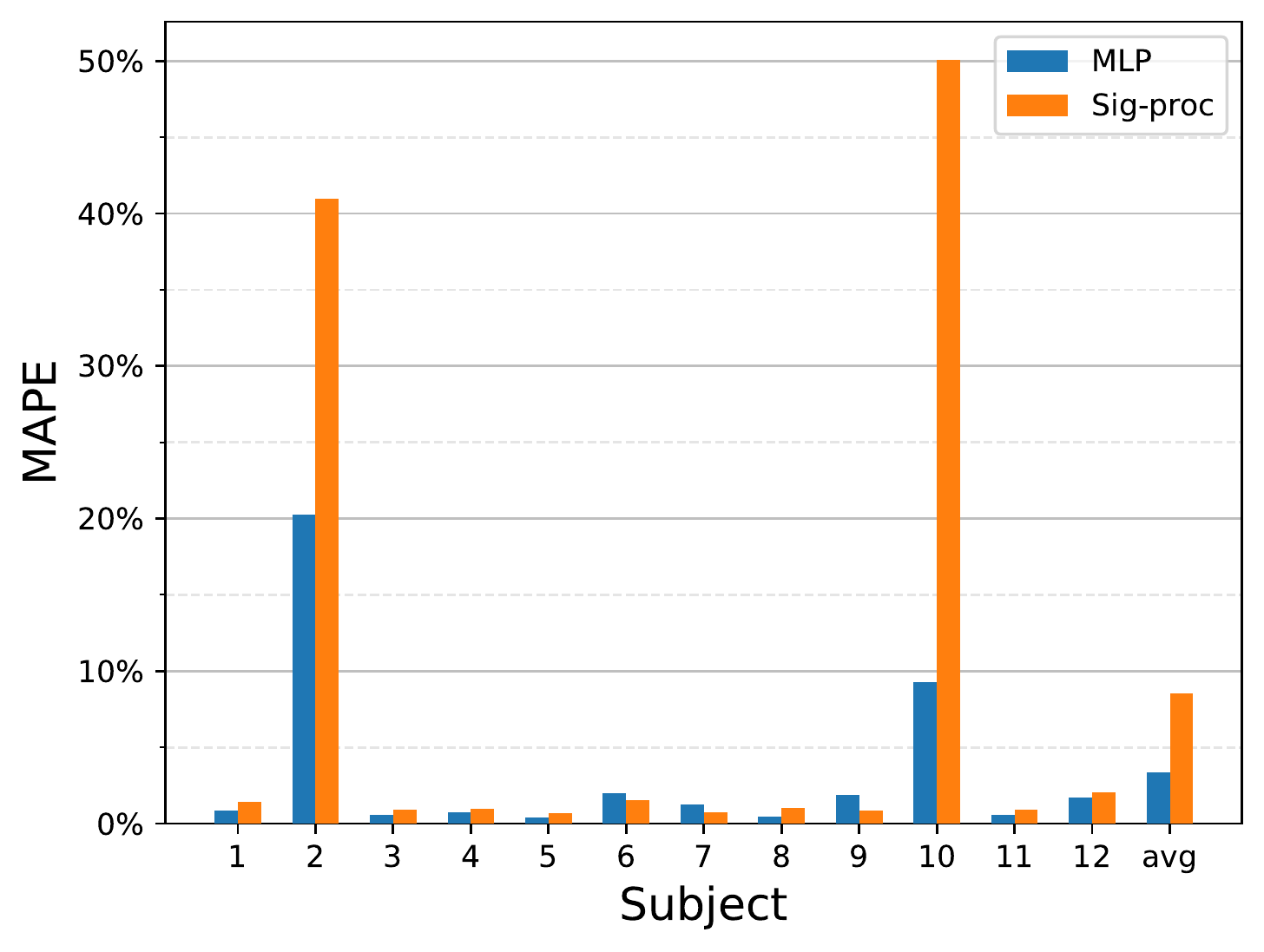}
  \caption{HR estimation MAPE for different subjects in the ISPC data set (MLP model's feature size is 2). "MLP" stands for our method that combines signal processing and MLP, and "Sig-proc" stands for the signal-processing-only method reproduced based on the ISPC paper~\cite{zhang2014troika}.}
  \label{fig:hr_mape_subject}
\end{figure}

\begin{figure}
  \centering
  \includegraphics[width=1\linewidth]{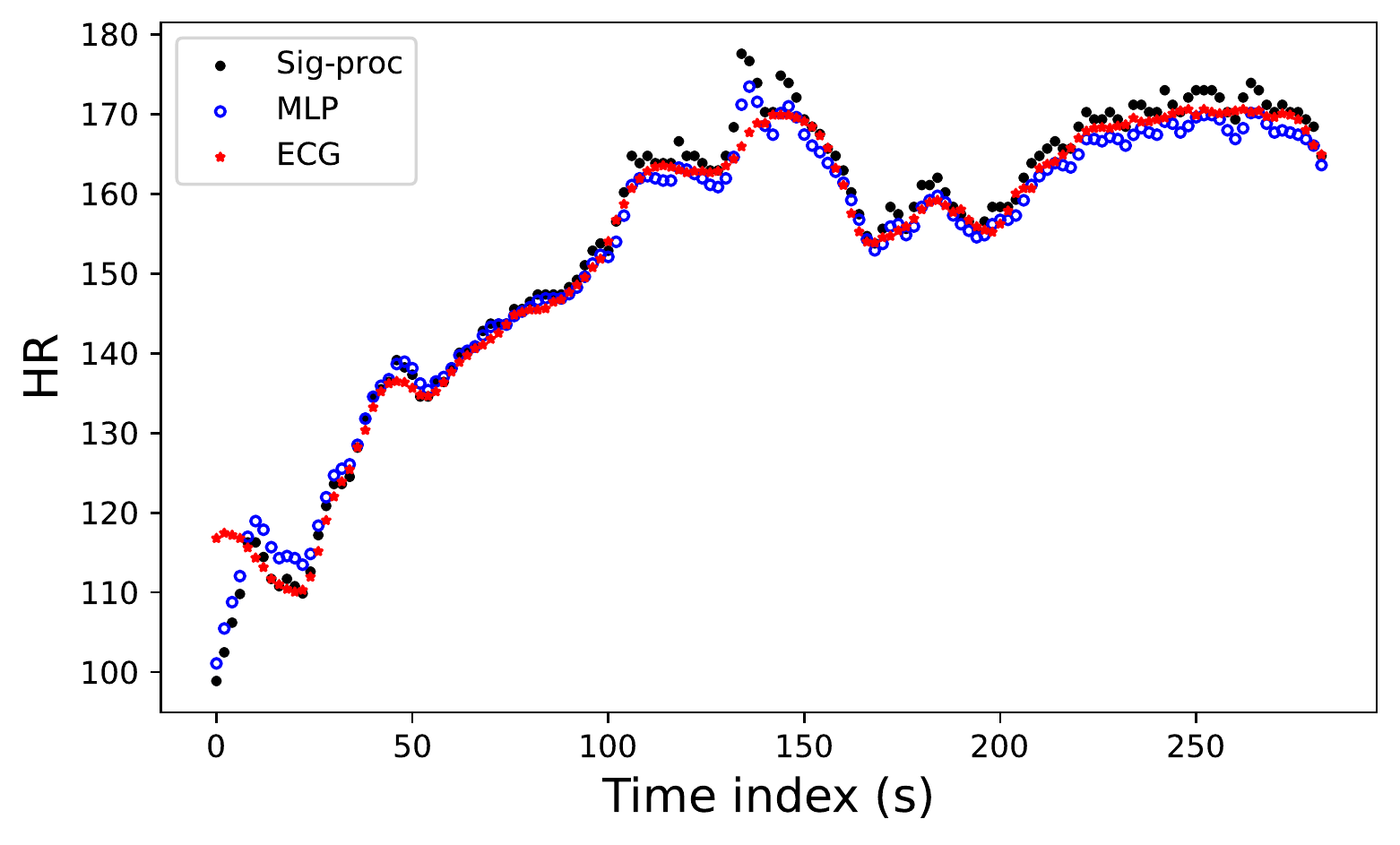}
  \caption{HR estimation trace for subject 11 in the ISPC dataset (the feature size of 2 is used for the MLP model).}
  \label{fig:hr_ispc_singleSubject}
\end{figure}

\begin{figure}
  \centering
  \includegraphics[width=1\linewidth]{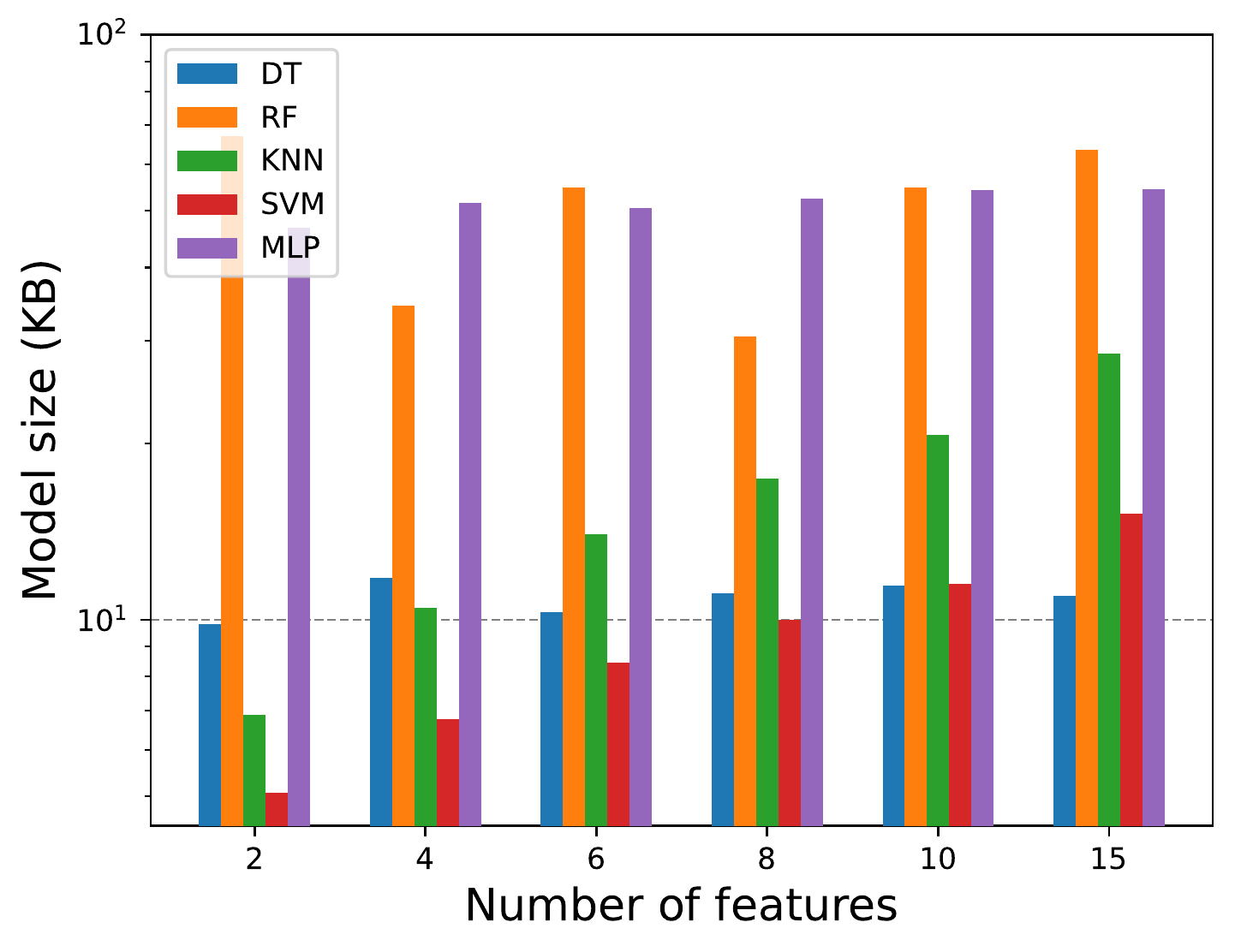}
  \caption{Model sizes for different ML models with different number of features when estimating HRs using the ISPC dataset.}
  \label{fig:hr_modelsize_ispc}
\end{figure}

\begin{figure}
  \centering
  \includegraphics[width=1\linewidth]{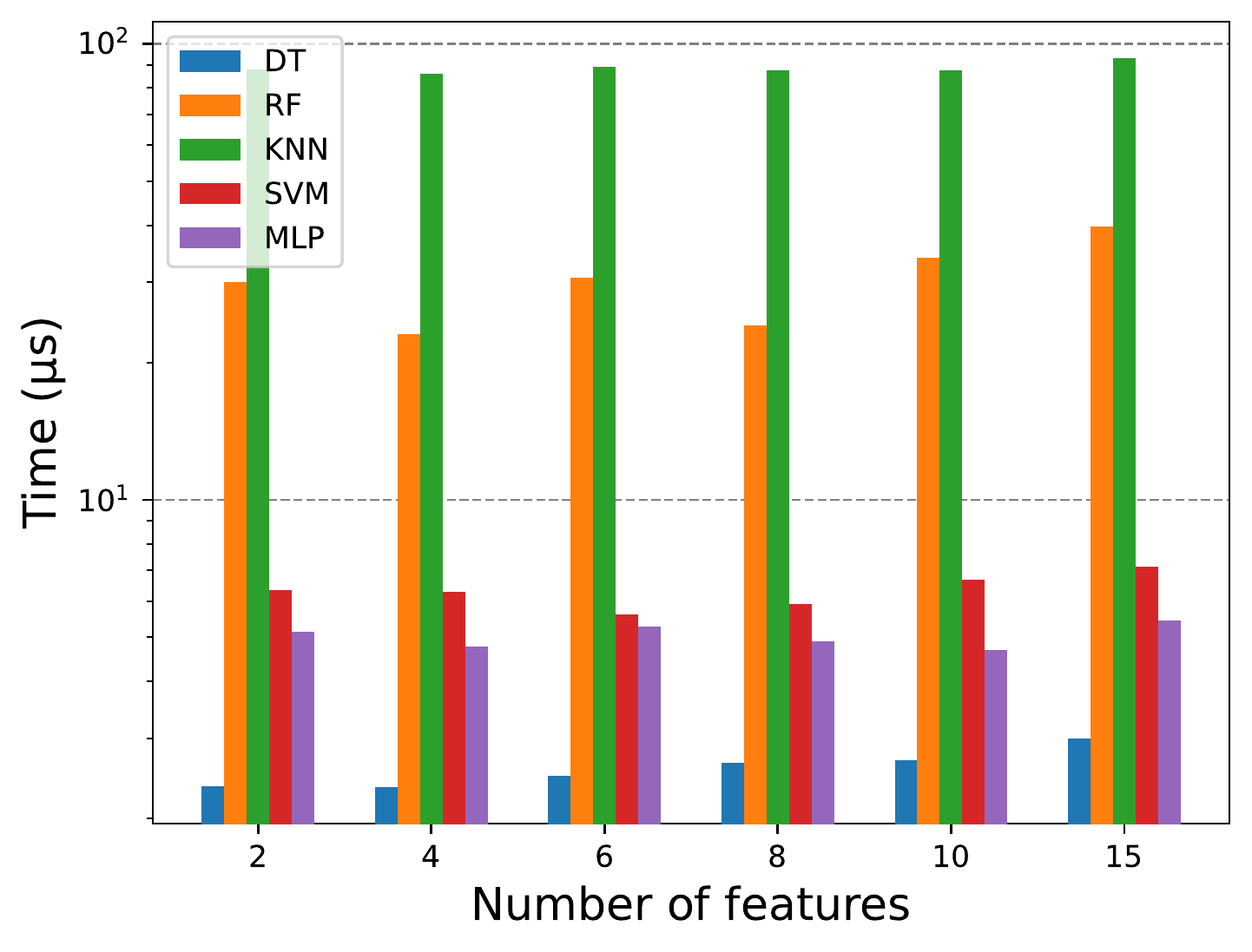}
  \caption{Model inference time for each HR reading with different ML models and number of features when estimating HRs using the ISPC dataset.}
  \label{fig:hr_time_ispc}
\end{figure}

\subsection{Heart Rate Estimation Results}

\subsubsection{HR Estimation Accuracy for ISPC Dataset}
Figure~\ref{fig:hr_mape_subject} gives the HR estimation errors for the ISPC dataset, including the MAPEs of our method that combines signal processing and MLP model, and the MAPEs of the signal-processing-only method. 
Note that, the MLP model here has two features (i.e., two PPG-HRs as features). 
As Figure~\ref{fig:hr_mape_subject} shows, our combined method usually had lower errors than the signal-processing-only method.
On average, our method had only 3.33\% error, which is 5.17\% lower than the signal-processing-only method. 
To further compare the accuracy differences, Figure~\ref{fig:hr_ispc_singleSubject} gives the HR estimations of our combined method, the signal-processing-only method, along with the ECG ground truth for subject 11. 
As Figure~\ref{fig:hr_ispc_singleSubject} shows, our HR estimations are more close to ECG readings, and less fluctuating than those from the signal-processing-only method, suggesting that the MLP model in our method could indeed reduce the random errors from the HRs generated by signal processing.

Figure~\ref{fig:hr_mape_subject} also indicates that, for most subjects, the HR estimation errors are below 2\%, suggesting that PPG-based HR monitoring can have high accuracy. 
The only two exceptions are subject 2 and subject 10, where the PPG signals were more noisy and random, likely due to worse MA effects during data collection. 
The same issue was also observed by other studies~\cite{zhang2015photoplethysmography}. 
Nonetheless, even for subjects 2 and 10, our combined method is still significantly more accurate than the signal-processing-only approach.

\subsubsection{ML Model Exploration for ISPC Dataset}
Besides MLP with two features, we also explored ML models and feature sizes to search for a configuration that can provide good accuracy with small model size and less inference time. 
Table~\ref{tab:hr_numofFeature_ispc} gives the MAPEs for different ML models and feature sizes (i.e., PPG-HR input sizes). Note that, the MAPEs in Table~\ref{tab:hr_numofFeature_ispc} is the average MAPEs for all 12 subjects for each configuration along with the standard deviation. 
As Table~\ref{tab:hr_numofFeature_ispc} shows, while all configurations have similar low errors, RF with 15 features has the lowest average MAPE at 2.62\%, followed closely by DT with 15 features at 2.76\%.

Figure~\ref{fig:hr_modelsize_ispc} gives the average model sizes for all subjects with different ML configurations, which shows that DT usually has smaller model sizes. 
While KNN and SVM have smaller model sizes when the number of features is small, due to their algorithmic philosophy, their model sizes grow rapidly when the number of features increases.

Figure~\ref{fig:hr_time_ispc} gives the average model inference time for all subjects with different ML configurations, which shows that DT usually has shorter inference time. The inference time did not increase much when the number of features increased from 2 to 15.
Considering the lower error from DT, these results suggest that DT with 15 features is a better configuration that provides both high accuracy, smaller model sizes, and short inference time.

\subsubsection{HR estimation Accuracy for Our Dataset}

\begin{figure}
  \centering
  \includegraphics[width=1\linewidth]{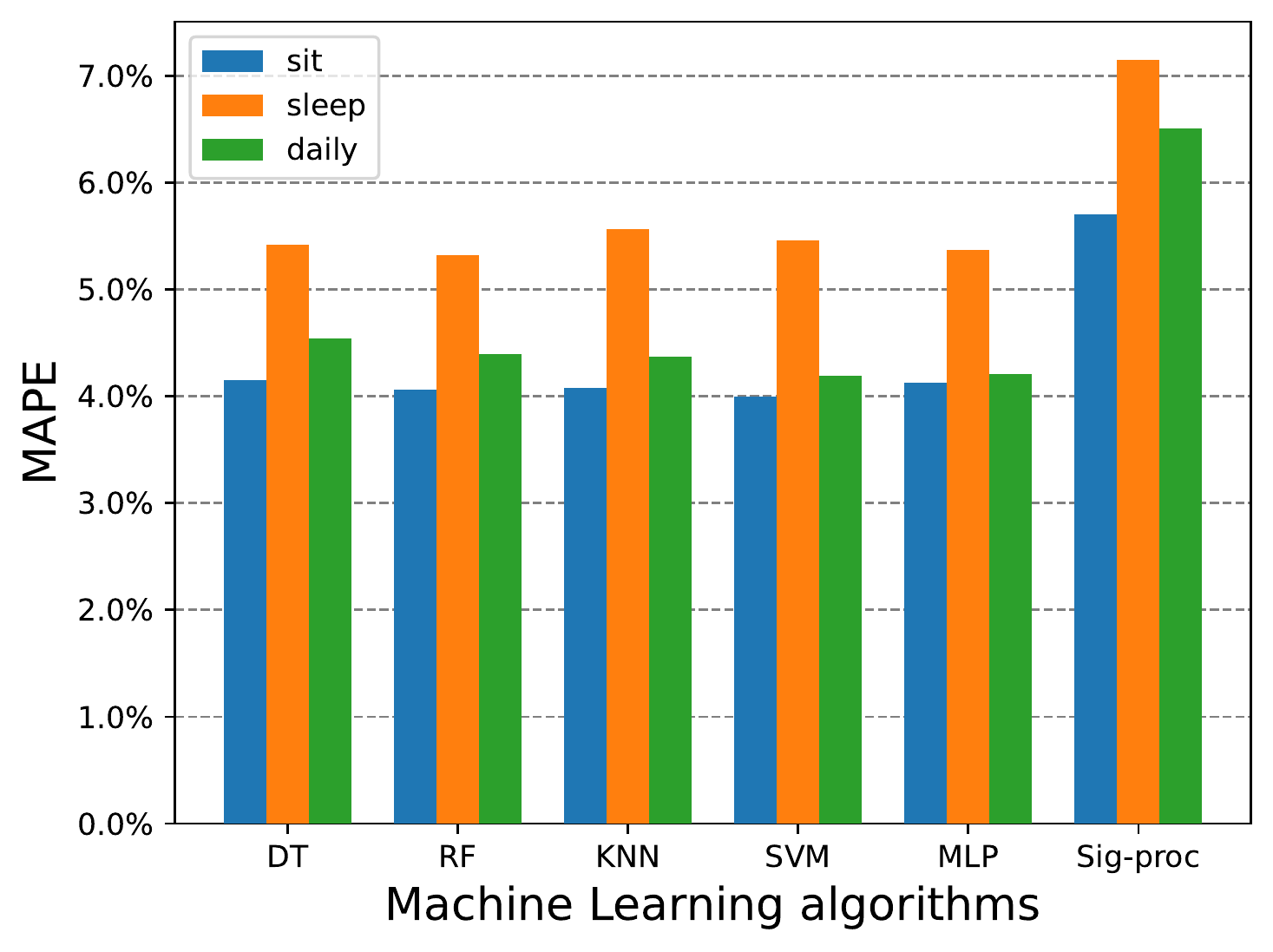}
  \caption{HR estimation MAPEs for different scenarios and ML algorithms using our data set (the feature size of 10 is used for ML models).}
  \label{fig:hr_alg_mape}
\end{figure}

\begin{figure}
  \centering
  \includegraphics[width=1\linewidth]{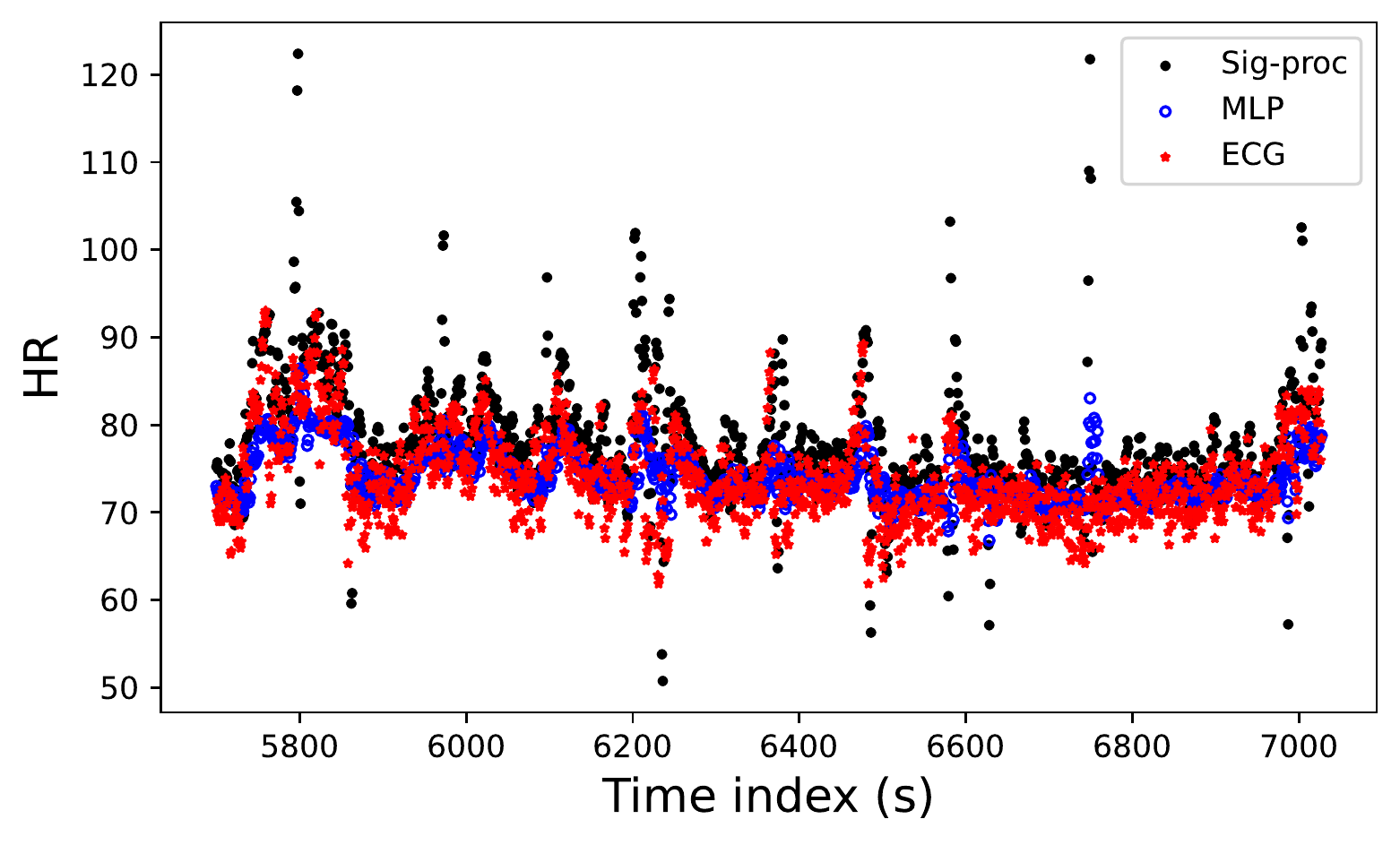}
  \caption{Partial testing set HR estimation trace for the daily activities scenario (MLP model's feature size is 10). "Sig-proc" stands for the signal-processing-only method described in Section \ref{sec:stage2}.}
  \label{fig:hr_ecgSensorPredicted}
\end{figure}

\begin{figure*}
  \centering
  \includegraphics[width=1\linewidth]{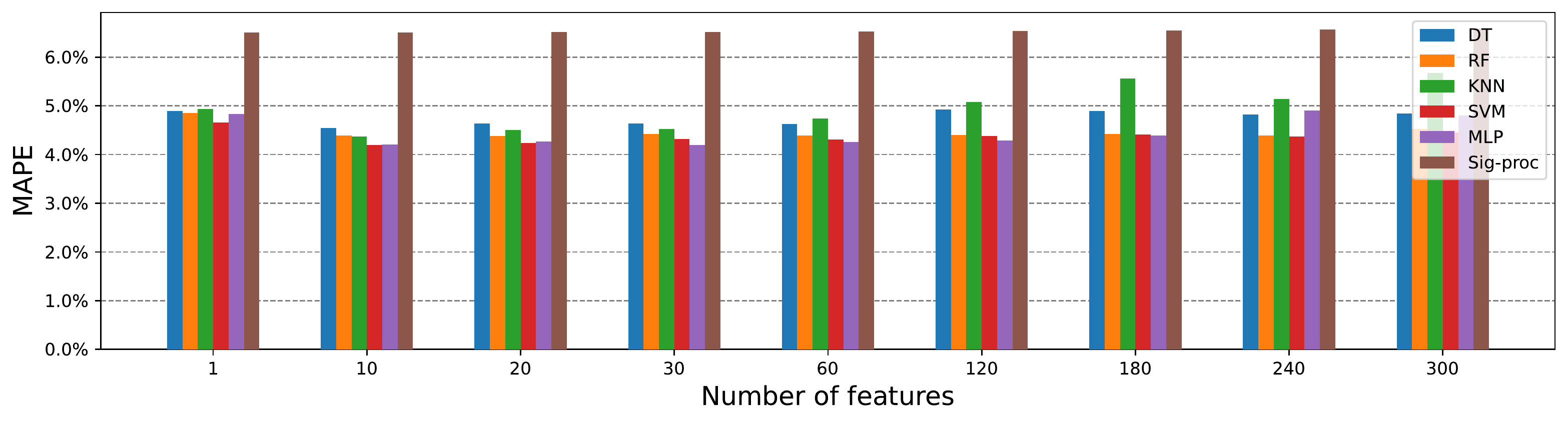}
  \caption{HR estimation MAPE for different number of features with our collected daily scenario data.} 
  \label{fig:hr_numofFeature}
\end{figure*}

\begin{figure*}
  \centering
  \includegraphics[width=1\linewidth]{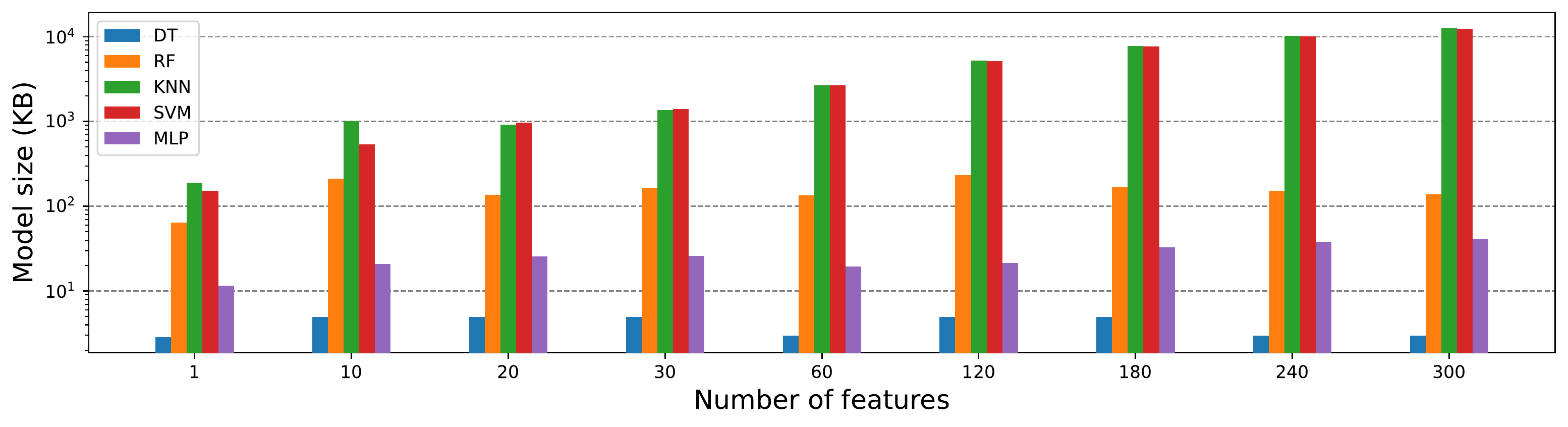}
  \caption{ML model size for different number of features when estimating HR using our collected daily scenario data.} 
  \label{fig:hr_modelsize}
\end{figure*}

\begin{figure*}
  \centering
  \includegraphics[width=1\linewidth]{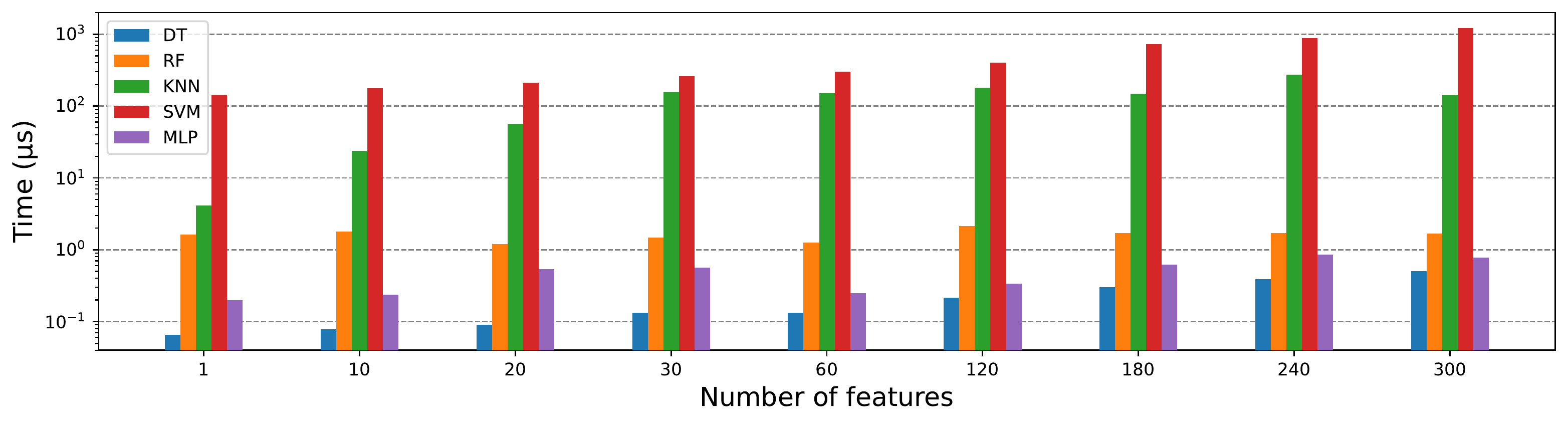}
  \caption{ML inference time for different number of features when estimating HR using our collected daily scenario data.} 
  \label{fig:hr_time}
\end{figure*}
Figure~\ref{fig:hr_alg_mape} gives the HR estimation error using our data set. 
Note that, the ML models here have 10 features (i.e., 10 PPG-HRs as features). 
As Figure~\ref{fig:hr_alg_mape} shows, our combined method with either DT, RF, KNN, SVM, and MLP models always has lower error than the signal-processing-only model (Sig-proc) for all activity scenarios. Moreover, the errors of our method are always below 6\%, showing that our method is highly accurate. 
Figure~\ref{fig:hr_ecgSensorPredicted} further gives the HR traces for our method with an MLP model, the signal-processing-only method, and the ECG ground truth. 
Similar to the ISPC evaluation results, HRs from our combined method in Figure~\ref{fig:hr_ecgSensorPredicted} are usually more close to ECG readings. Our HR estimations are also less fluctuating than those from the signal-processing-only method, suggesting that our method could reduce the random noises in the PPG signals when the PPG signal is only at 25Hz.

\subsubsection{ML Model Exploration for Our Dataset}

Again, we also explored ML models and feature sizes using our data set to search for a configuration with good accuracy, small model size, and fast inference time. 
Since our data duration is longer than ISPC data, we took advantage of this and explored up to 300 features to further investigate how much historical data is preferable as ML input.
Figure~\ref{fig:hr_numofFeature} gives the MAPEs for different ML models and feature sizes (i.e., PPG-HR input sizes) for the daily activity scenario. We can see from Figure~\ref{fig:hr_numofFeature}, that the errors for all types of ML models are similarly low when there are more than 10 features, suggesting that 10 to 20 features could be enough for our data set.

Figure~\ref{fig:hr_modelsize} gives the model sizes of different ML configurations, which shows that DT models are consistently the smallest. 
Figure~\ref{fig:hr_time} provides the model inference time with different ML configurations, which reveals that DT models are consistently the fastest. 
All the inference times increased slightly when the number of features increased from 1 to 300, due to more features and bigger models. 
Considering the lower error from DT, these results suggest that DT with 10 to 20 features is a better configuration that provides high accuracy with a small model size and fast inference time.

\section{Conclusion}
\label{sec:Conclusion}


While Photoplethysmography (PPG) sensors could provide accurate HR estimations, applying these sensors to embedded devices is still challenging due to the high-frequency PPG sampling, which has high power consumption, and the complex machine-learning models, which are too computational-intensive and large for small embedded devices. 
In this paper, we showed that combining signal processing and ML could significantly reduce PPG sampling frequency while providing high HR estimation accuracy. The experimental results showed that our method that combined signal processing and ML had only $5\%$ error for HR estimation using  low-frequency PPG data. This combination also reduces the ML model's feature size, leading to smaller models.
Additionally, we also conducted a comprehensive analysis of different ML models and feature sizes to compare their accuracy, model size, and inference time. Our analysis showed that DT models with 10 to 20 features usually have good accuracy while being several magnitude smaller in model sizes and faster in inference time.



\bibliographystyle{abbrv}
\bibliography{bib-references}

\end{document}